\documentclass[graybox]{svmult}
\usepackage{graphicx}
\usepackage{mathptmx}       
\usepackage{helvet}         
\usepackage{courier}        
\usepackage{type1cm}        
\usepackage{makeidx}         
\usepackage{graphicx}        
\usepackage{multicol}        
\usepackage[bottom]{footmisc}

\makeindex             

\begin{document}

\title*{Materials that make robots smart}
\titlerunning{Materials that make robots smart}  
%
\author{Nikolaus Correll$^{1,2}$ \and Christoffer Heckman$^1$
}
\authorrunning{Correll and Heckman} 
%

\institute{Nikolaus Correll \at University of Colorado at Boulder, \email{ncorrell@colorado.edu}
\and Christoffer Heckman \at University of Colorado at Boulder \email{christoffer.heckman@colorado.edu}}

\tocauthor{Nikolaus Correll and Christoffer Heckman}
\institute{University of Colorado at Boulder\\
$^1$Department of Computer Science\\
$^2$Material Science Engineering}

\maketitle              
\abstract*{
We posit that embodied artificial intelligence is not only a computational, but
also a materials problem. While the importance of material and structural
properties in the control loop are well understood, materials can take an
active role during control by tight integration of sensors, actuators,
computation and communication.  We envision such materials to abstract
functionality, therefore making the construction of intelligent robots more
straightforward and robust. For example, robots could be made of bones that
measure load, muscles that move, skin that provides the robot with information
about the kind and location of tactile sensations ranging from pressure, to
texture and damage, eyes that extract high-level information, and brain
material that provides computation in a scalable manner. Such materials will
not resemble any existing engineered materials, but rather the heterogeneous
components out of which their natural counterparts are made. We describe the
state-of-the-art in so-called ``robotic materials,'' their opportunities for
revolutionizing applications ranging from manipulation to autonomous driving,
and open challenges the robotics community needs to address in collaboration
with allies, such as wireless sensor network researchers and polymer scientists.}

\abstract{ We posit that embodied artificial intelligence is not only a
computational, but also a materials problem. While the importance of material
and structural properties in the control loop are well understood, materials
can take an active role during control by tight integration of sensors,
actuators, computation and communication.  We envision such materials to
abstract functionality, therefore making the construction of intelligent robots
more straightforward and robust. For example, robots could be made of bones
that measure load, muscles that move, skin that provides the robot with
information about the kind and location of tactile sensations ranging from
pressure, to texture and damage, eyes that extract high-level information, and
brain material that provides computation in a scalable manner. Such materials
will not resemble any existing engineered materials, but rather the
heterogeneous components out of which their natural counterparts are made. We
describe the state-of-the-art in so-called ``robotic materials,'' their
opportunities for revolutionizing applications ranging from manipulation to
autonomous driving, and open challenges the robotics community needs to address
in collaboration with allies, such as wireless sensor network researchers and polymer
scientists.}

\section{Introduction}
\begin{figure}[!htb]
\vspace{-20px}
\includegraphics[width=\columnwidth]{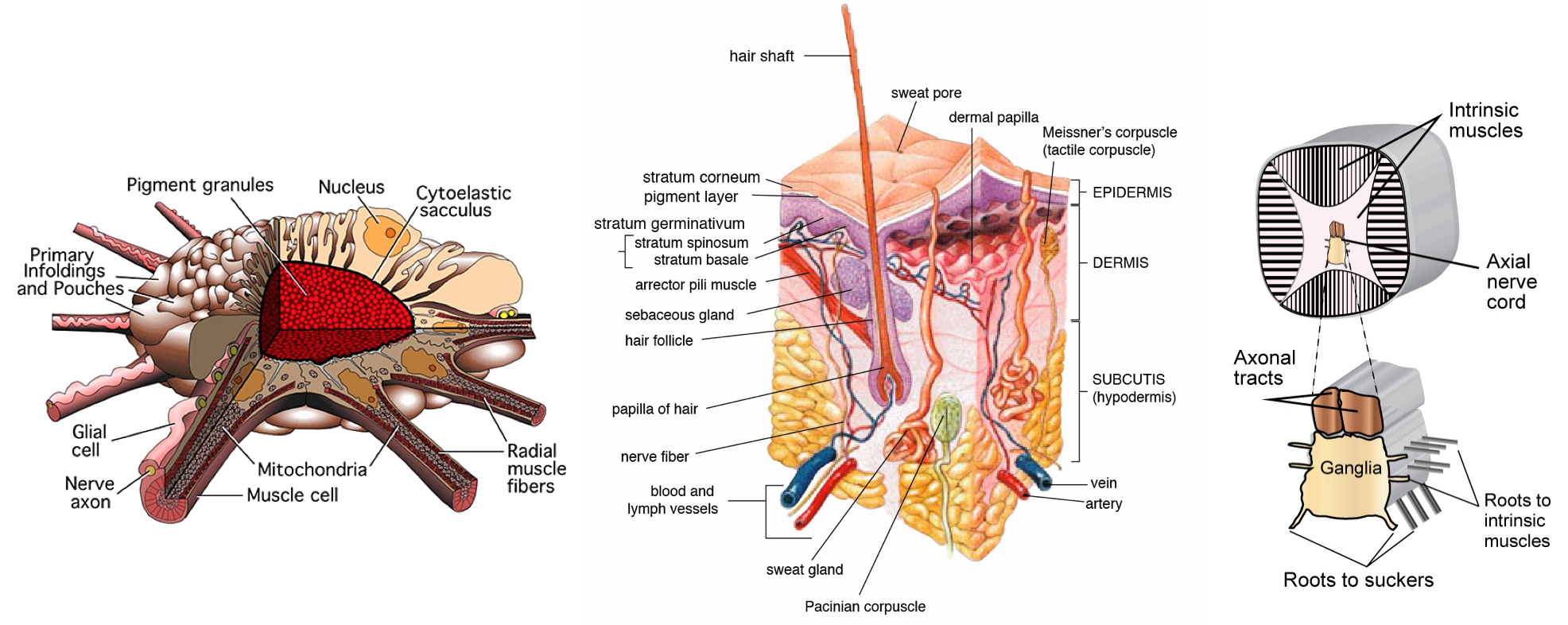}
\caption{Biological tissue that tightly integrates sensing, actuation, computation and communication. From left to right: chromatophore in an octopus skin (\copyright Springer Verlag, from \cite{cloney1968ultrastructure}), human skin (\copyright Public Domain), octopus suckers. Computation and communication are implemented by a nervous system, power is provided by a vascular system. \label{fig:bio}}
\end{figure}

The impressive functionality of natural systems such as the camouflage skin of
a cuttlefish, morphing wings of an eagle, structural adaptation of a mammalian
bone or the Banyan tree, or the many sensor modalities in the human skin are
enabled by tight integration of sensing, actuation, computation and
communication in these systems \cite{mcevoy15}. Figure \ref{fig:bio} shows
schematic drawings of some of these natural systems, illustrating how function
arises from sensors, actuators, nerves and vascular systems being co-located. 

\begin{figure}[!htb]
\vspace{-15px}
\includegraphics[width=\columnwidth]{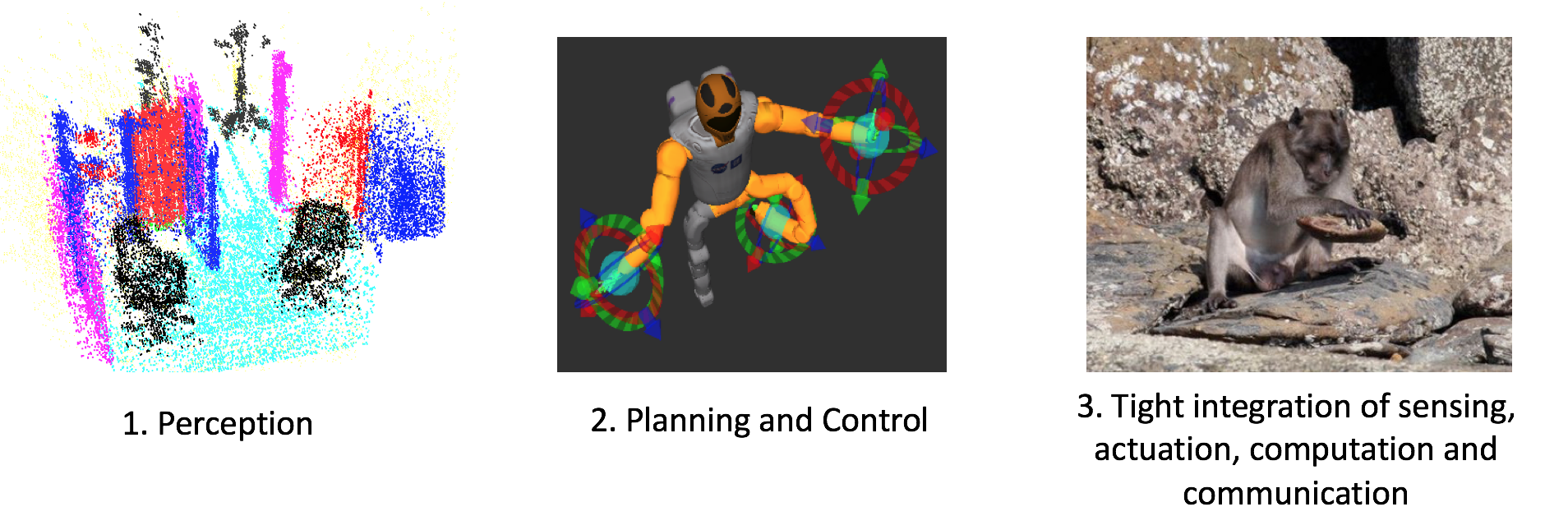}
\caption{Most robotic systems clearly separate between perception and control. Here, the role of the material is to make the transition from a computational model to real-world dynamics as smooth as possible. This is not the case in biological systems that are sensor rich, dynamic and over-actuated.\label{fig:robotics}}
\end{figure}

This is fundamentally different from how we construct robotic systems, which consists of homogeneous, hierarchical components such as structures, gears, links and joints that are interfaced by sensing and computer systems, but lack the tight integration of sensing and computation that biological systems exhibit (Figure \ref{fig:robotics}). 

This paradigm has begun to change with the emergence of soft robotics and multi-material manufacturing techniques \cite{cutkosky2009design}, which has led to fully self-contained robotic systems \cite{bartlett20153d} and components \cite{asbeck2006scaling,kim2006isprawl}
that blur the distinction between a robot and a material
\cite{correll2014soft}. While soft robotics provides radically novel ways for
locomotion \cite{marchese2014autonomous,tolley2014untethered} and manipulation
\cite{farrow2016morphological}, this field has led to new manufacturing
approaches that allow us to get much closer to the integrated nature of
biological systems \cite{polygerinossoft,cutkosky2009design}. 

We believe it is this integration that will allow us to create robotic
components that are in some sense and by themselves autonomous, thereby
facilitating the creation of more complex and adaptable autonomous robotic
systems. We argue that such components, which necessarily implement design
trade-offs and are therefore more suitable for one robot design than another,
reach their true potential once they are available in the form of a
\emph{material}.

A definition of ``material'' that extends to these devices is a critical
element to this newly developing field. This quandary permeated the halls of
the first ``Workshop on Robotic Materials'', which was held March 10--12, 2017,
in Boulder, CO. The consensus reached by that workshop were that materials have
the properties of:

\begin{enumerate}
  \item \textbf{Functionality independent of size}, i.e., performance is unchanged when cut in half (up to a reasonable discretization);
  \item \textbf{Self-similarlity and bulk reconfigurability}, i.e., consisting of homogeneous elements that can be arranged in either an amorphous or a discrete, grid-like fashion; and
  \item \textbf{Robustness}, i.e.\ the material does not lose its capabilities should failure of any constituent elements occur.
\end{enumerate}

These properties align well with the contemporary material science perspective
\cite{brostow2016materials}. Here,``robotic materials'' are composites with
structural, sensing, actuation, and computational ``phases.'' Such phases can
be either dispersed, such as in composites that are reinforced with granular
particles, or placed in an anisotropic manner, such as in fiber reinforced
composites. While providing the material with improved functionality,
additional phases typically introduce challenges in terms of structural
integrity and manufacturing at the interfacial bond between the matrix and the
dispersed phases. Here, robotic materials pose hard problems by requiring the
integration of hard elements into soft materials, as well as materials and
devices that are not designed to create strong bonds with others.

Once the conditions that define a material are met and manufacturing challenges
are solved, such materials could be manufactured and distributed in volumetric
units and used to enhance the functionality of a robotic system. Section
\ref{sec:soa} describes a series of such systems, not necessarily all with
robotic applications, that emphasize the distributed algorithms, material
science, communication, power, and manufacturing challenges. Section
\ref{sec:vision} then describes what kind of materials we envision to transform
how we make robots.

\section{State-of-the-art}\label{sec:soa}
\begin{figure*}
\includegraphics[height=1.335in]{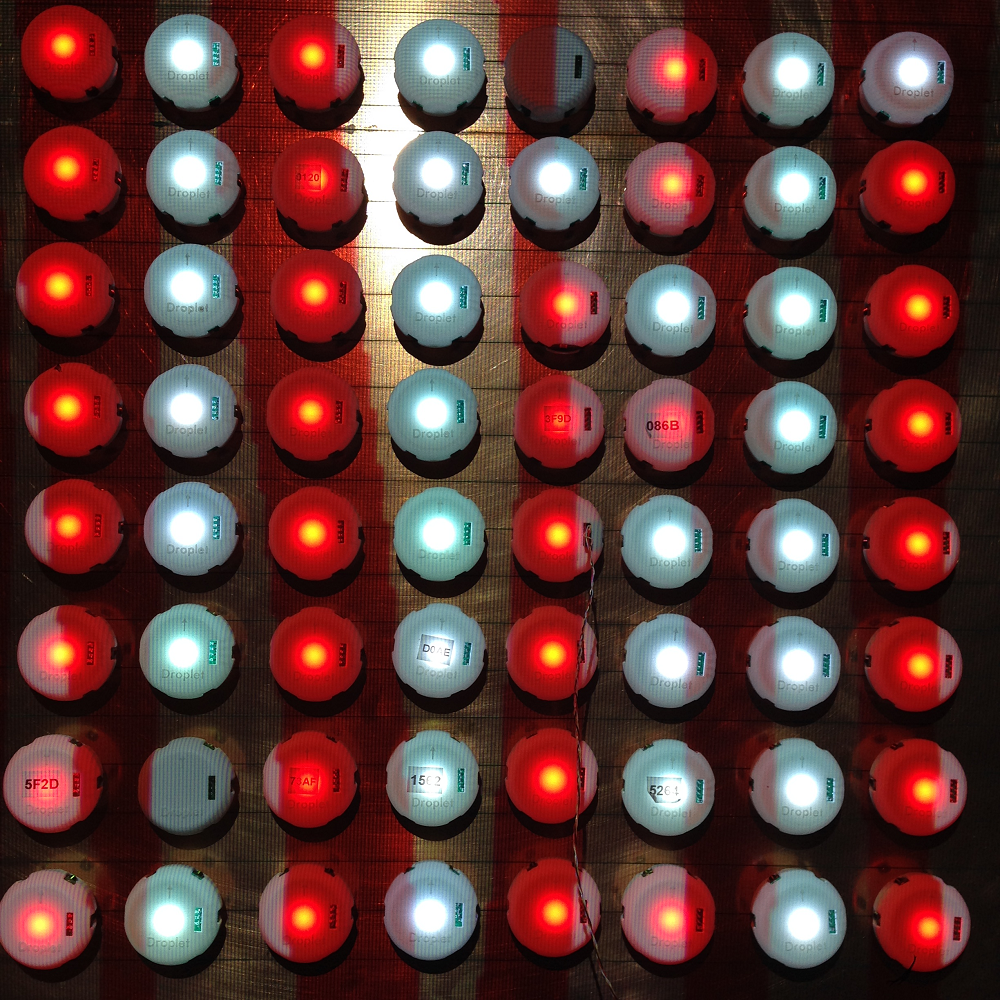}
\includegraphics[height=1.335in]{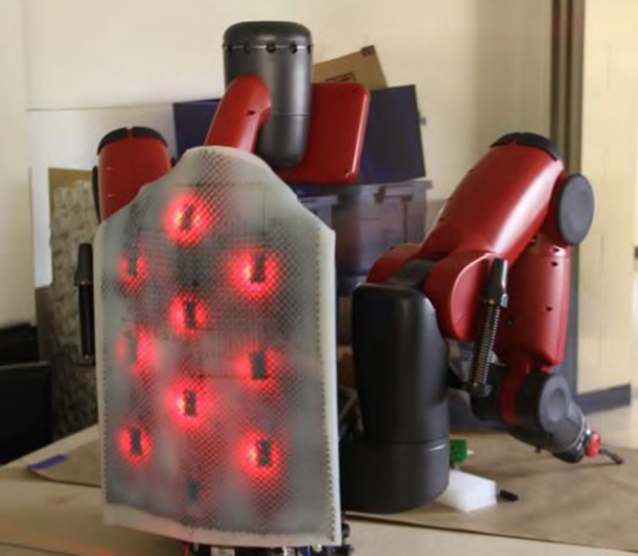}
\includegraphics[height=1.335in]{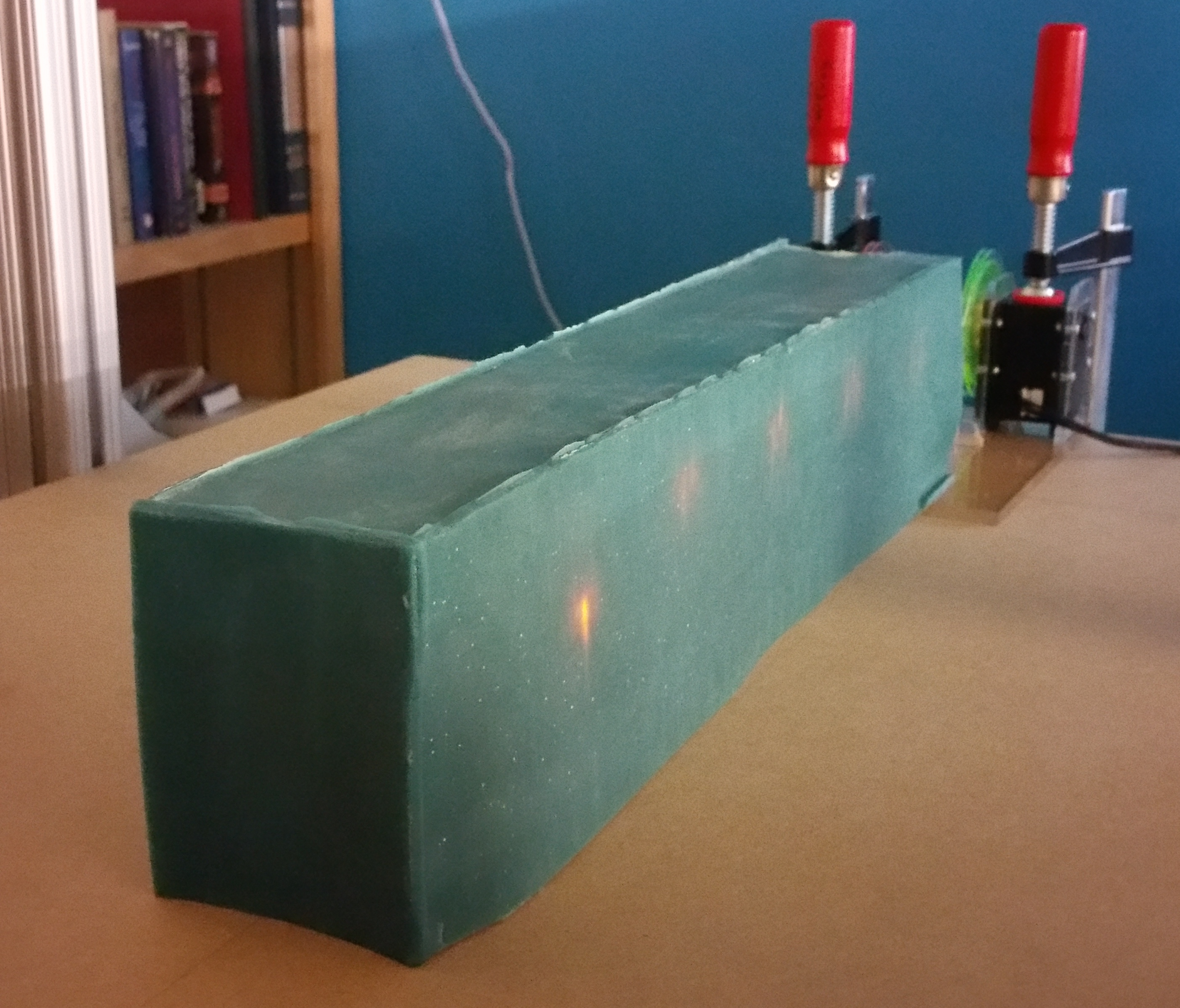}\\
\includegraphics[height=1.33in]{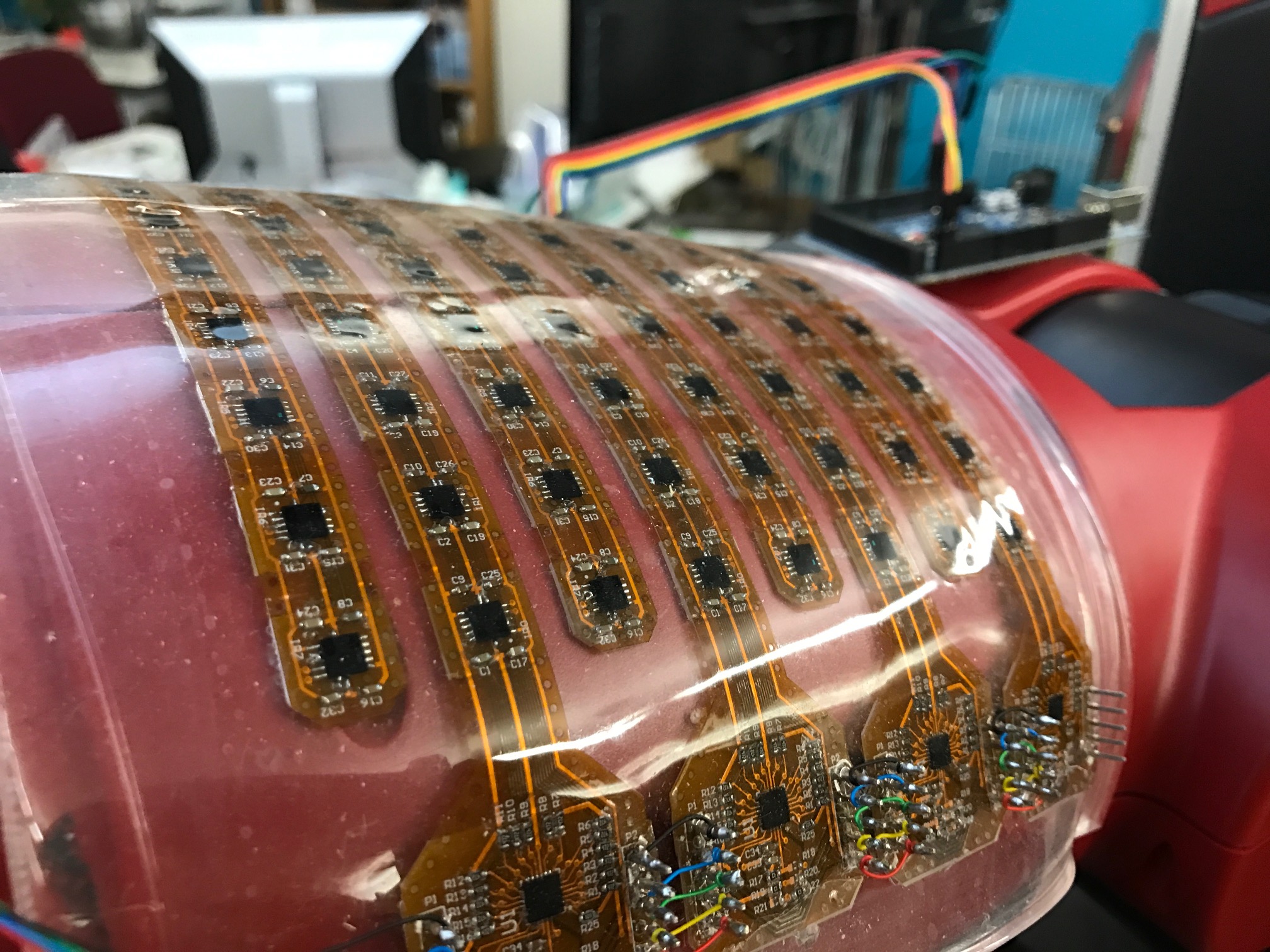}
\includegraphics[height=1.33in]{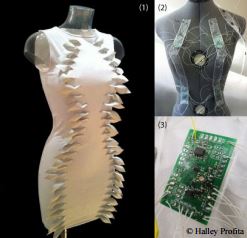}
\includegraphics[height=1.33in]{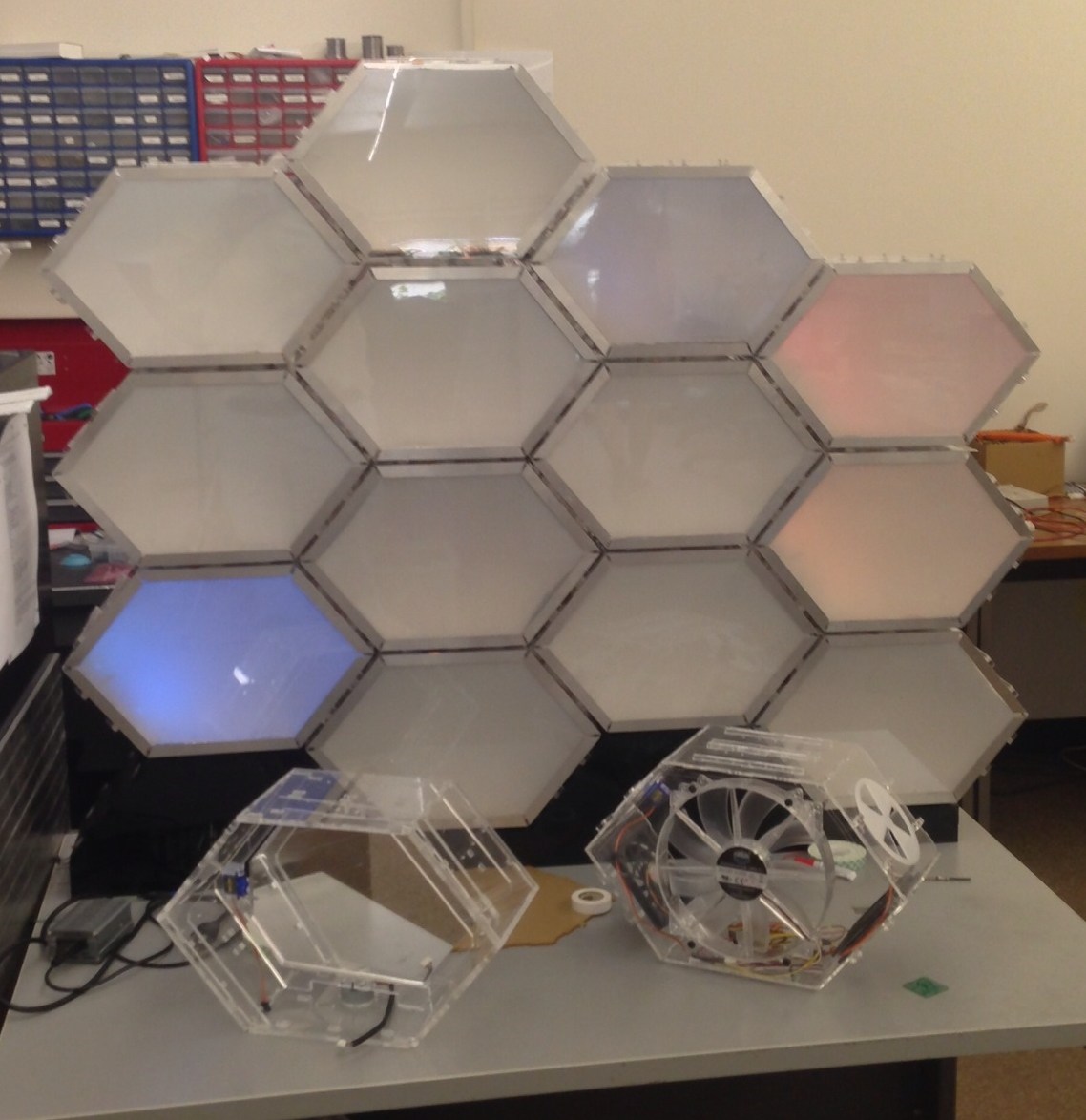}

\caption{From top-left to bottom-right: active camouflage \cite{li2016}, texture-detecting skin \cite{hughes2015texture}, a shape-changing beam \cite{mcevoy2016shape},  gesture-detecting skin \cite{hughes2015detecting}, a sound localizing dress \cite{profita2015flutter}, and an interactive fa\'cade \cite{hosseinmardi2015distributed}.\label{fig:examples} }
\end{figure*}

Figure \ref{fig:examples} shows a series of systems which, although crude,
adhere to the necessary conditions for making them a material as laid out in
the definition of materials enunciated above.

The camouflage system (top-left) consists of a swarm of ``Droplets'' miniature robots, which perform consensus and distributed pattern generation algorithms to match dominant patterns they perceive in the environment \cite{li2016}. As the algorithm is fully local, the system works no matter what the shape or size of the arrangement is, making an integration of color-changing particles into rubber sheets conceivable.

The texture sensing skin (top-center) is able to localize objects as well as differentiate up to 15 different textures by measuring the vibrations that are induced when rubbing against the skin \cite{hughes2015texture}. The algorithm is fully distributed, and a network of microcontrollers has been integrated into a rubber skin. 

The shape-changing beam (top-right) consists of six identical variable stiffness elements \cite{mcevoy2016shape}, which can calculate the required stiffness to reach an arbitrary shape after bending in a fully distributed way \cite{mcevoy2016distributed}. The proposed algorithm scales linearly with the length of the beam, albeit allows for instantaneous motion, reducing the computation to a constant time operation with respect to the system's motion. 

 The gesture-detecting skin (bottom-left) consists of 8$\times$8 proximity sensors integrated into transparent polymer and is capable to distinguish between a variety of social touch gestures \cite{hughes2015detecting}. Sensor information is accessed using a hierarchical bus system and processed for an area of 8$\times$8 by a deep-learning network that fits onto a microcontroller. This system fulfills the material requirements above with the shown patch as individual unit, making a large deployment conceivable in which high-level gesture information are communicated in a hop-by-hop fashion.
 
The dress (bottom-center) is the prototype of a smart fabric that uses a regular arrangement of computing elements to triangulate the direction incoming sounds \cite{profita2015flutter}. 

The facade system in the lower-right is created from identical building blocks that are arranged in a regular structure and are each equipped with the ability to sense a human hand and change their color \cite{farrow2014gesture}. Each block shares power and local communication with its neighbors. By locally exchanging information about touch events, the system is able to recognize a series of letters drawn across its surface \cite{hosseinmardi2015distributed}, functions no matter how many elements the surface has, and is able to interpolate across broken elements.

\section{Vision and Challenges}\label{sec:vision}
The examples above demonstrate the overall feasibility of obtaining complex
signal processing and control ability from fully distributed systems. All of
these materials could be made with smaller components and integrated at higher
density, and all of them rely only on local information, making them scalable.
Some of the presented materials have already been created with robotic
applications in mind. For example, skins that localize and recognize gestures
and textures could dramatically facilitate the robot design problem in highly
asymmetric ways. If embedded into an end effector, grasp selection and reactive
control would be available based on local information from the skin
\cite{patel2017improving} rather than through exteroceptive sensors perceiving geometry. Not only does
local processing take the burden of a central processing unit, but also
directly addresses the problems of registering the location of individual
sensors on the robotic skin and routing the information. Similarly, a shape
changing beam would allow to control the structure of a robotic system such as
running and walking robots, exo-skeletons, but also airfoils of aircraft or
autonomous vehicles.

Once we are ready to delegate to a material the functionality that has previously only been available at the device level, many other applications become conceivable. For example, smart rubber embedded with light-emitting computational elements coated over a robot body would allow it to camouflage, act as a display for arbitrary information, or indicate material fatigue.
Embedding a polymer with proximity sensors could not only create robotic skins, but smart wheels that can measure the ground they are on \cite{hughes2017} or the tire profile from the inside to detect skid, and turn any solid into an input device that is aware of its surroundings. Using accelerometers and gyroscopes that can compute and communicate with their local neighbors would allow a material to ascertain proprioceptive state with unprecedented spatial resolution. 
Functionality can also come from the polymer itself. For example, polymers might contract, expand, or change their stiffness, viscosity or color when locally activated by light, electric current or magnetic fields \cite{mcevoy15}, shape-changing structures, or tires that adapt their profile to driving conditions. 

Interestingly, many of the key technologies to enable these materials already exist.  Computational elements have been dramatically reduced in size, making the operation of thousands of wireless devices in close proximity feasible. This brings us close to the original vision of distributed MEMS and ``smart dust'' \cite{berlin1997distributed} and challenges specific to the wireless networking community are described in \cite{sensys}.

Given a small enough computational element with wireless communication
\cite{pannuto2014networked}, many conceivable robotic materials could be
produced using existing manufacturing techniques for composite materials
such as vacuum forming, shape deposition manufacturing and overmolding. A key
challenge of such highly-functioning materials is the wide range of length
scales, ranging from microscale sensors and computational elements to the
meter-scale, to which the final material must extend. It is this scaling
property that leads most conventional manufacturing techniques to their
limitations. One possible solution to this problem is to assemble robotic
materials using autonomous robots or multi-material 3D printers
\cite{o2017review}. Given the ability to provide power wirelessly
\cite{agbinya2015wireless} and to localize the computational elements inside
the structure after manufacturing, either using external fields or using
self-localization technologies \cite{moffo2016relative}, it is also conceivable
that it may soon be possible to simply mix small pellets that sense,
communicate and compute and trigger actuation into a liquid material and let it
cure in the desired shape.

Reaching the desired homogeneous distribution of dispersed particles in a
polymer matrix is a problem that is well studied in the composite materials
community for nanoscale elements \cite{xie2005dispersion}, but more difficult
to achieve for objects at the millimeter scale. Robotic materials will also
benefit from advances in polymer themselves ranging from electro-active polymer
muscles \cite{keplinger2013stretchable} with improved performance to a wide
range of smart polymers that can double as sensors or actuators
\cite{bauer201425th}. 

Challenges lie not only in developing the different components of robotic
materials, however, but also their interaction. Integrating sensors, for
example, offers the opportunity of ``morphological computation''
\cite{pfeifer2006morphological}, that is the placement of sensors in a way that
minimizes the computations required to process them. Similarly, choosing
material properties carefully will affect the way the systems into which they
are integrated are controlled, a challenge unto itself.

It is clear that the vision of robotic material requires the close
collaboration of researchers in disparate fields like wireless networking,
robotics, and chemistry. Yet, bringing those groups together to create systems
that go beyond individual contributions at device level remains a major
challenge. In particular, isolated contributions from any individual discipline
are often difficult to transfer as challenges of system integration have been
neglected, possibly leading to designs that are fundamentally infeasible.   

\section{Conclusion}
Materials that make robots smart have the potential to dramatically simplify robot design by off-loading signal processing and control into the material, thereby abstracting some traditionally high-level functions. Creating such materials poses many challenges in distributed algorithms, manufacturing, as well as platform technologies such as wireless power transfer, directional RF communication, localization techniques, and polymers with interesting capabilities that can enhance sensing and actuation. 

Albeit somewhat functional, current robotic materials are too crude to be practical and will benefit from the continued development of these techniques and an improved understanding of how to integrate them. It is ultimately the robotic applications enabled by such materials that need to be compelling, leading then to interdisciplinary efforts that combine wireless sensor networking experts, material scientists and roboticists. From these first conversations, materials that enable perception by performing distributed computation seem to be more immediate candidates than materials that perform actuation. While the application of such materials are not limited to robotics, it is the robotics community that is well-positioned to appreciate the challenges that span from material to computer science and contribute to the multi-modal problem solving required.

\section*{Acknowledgments} This work was sponsored by ARO under grant number 
W911NF-16-1-0476, program manager S. Stanton, by AFOSR, program manager B.
``Les'' Lee, and DARPA award no.\ N65236--16--1--1000. We are grateful for this support.
\bibliographystyle{abbrv}
\bibliography{paper,isrr} 

\end{document}